\pdfoutput=1

\documentclass[11pt]{article}

\usepackage[final]{acl}

\usepackage{times}
\usepackage{latexsym}

\usepackage[T1]{fontenc}

\usepackage[utf8]{inputenc}

\usepackage{microtype}

\usepackage{inconsolata}

\usepackage{graphicx}
\usepackage{hyperref}       %
\usepackage{url}            %
\usepackage{booktabs}       %
\usepackage{amsfonts}       %
\usepackage{nicefrac}       %
\usepackage{microtype}      %
\usepackage{xcolor}         %
\usepackage{amsmath}
\usepackage{caption}
\usepackage{xspace}
\usepackage{multirow} 
\usepackage{enumitem}
\usepackage{colortbl}

\newcommand{\ours}{\textsc{STEM-PoM}\xspace}

\title{
\ours: Evaluating Language Models Math-Symbol\\ Reasoning in Document Parsing
}

\makeatletter
    \def\@fnsymbol#1{\ensuremath{\ifcase#1\or \dagger\or \ddagger\or
       \mathsection\or \mathparagraph\or \|\or **\or \dagger\dagger
       \or \ddagger\ddagger \else\@ctrerr\fi}}
\makeatother

\author{
Jiaru Zou\textsuperscript{1}\thanks{Corresponding authors}, 
Qing Wang\textsuperscript{1}, 
Pratyush Thakur\textsuperscript{1}, 
Nickvash Kani\textsuperscript{1}\footnotemark[1]
\\
\textsuperscript{1} University of Illinois Urbana-Champaign \\
\texttt{\{\href{mailto:jiaruz2@illinois.edu}{jiaruz2}, \href{mailto:kani@illinois.edu}{kani}\}@illinois.edu}
}

\begin{document}
\maketitle

\begin{abstract}
Advances in large language models (LLMs) have spurred research into enhancing their reasoning capabilities, particularly in math-rich STEM (Science, Technology, Engineering, and Mathematics) documents.
While LLMs can generate equations or solve math-related queries, their ability to fully understand and interpret abstract mathematical symbols in long, math-rich documents remains limited. In this paper, we introduce \ours, a comprehensive benchmark dataset designed to evaluate LLMs' reasoning abilities on math symbols within contextual scientific text. The dataset, sourced from real-world ArXiv documents, contains over 2K math symbols classified as main attributes of variables, constants, operators, and unit descriptors, with additional sub-attributes including scalar/vector/matrix for variables and local/global/discipline-specific labels for both constants and operators. 
Our extensive experiments demonstrate that state-of-the-art LLMs achieve an average accuracy of 20--60\% under in-context learning and 50--60\% with fine-tuning, highlighting a substantial gap in their ability to classify mathematical symbols. By improving LLMs' mathematical symbol classification, \ours further enhances models' downstream mathematical reasoning capabilities. The code and data are available at \href{https://github.com/jiaruzouu/STEM-PoM}{https://github.com/jiaruzouu/STEM-PoM}.

\end{abstract}

\section{Introduction}
\label{sec:introduction}

\begin{figure}[!t]
    \centering
    \includegraphics[width=\linewidth]{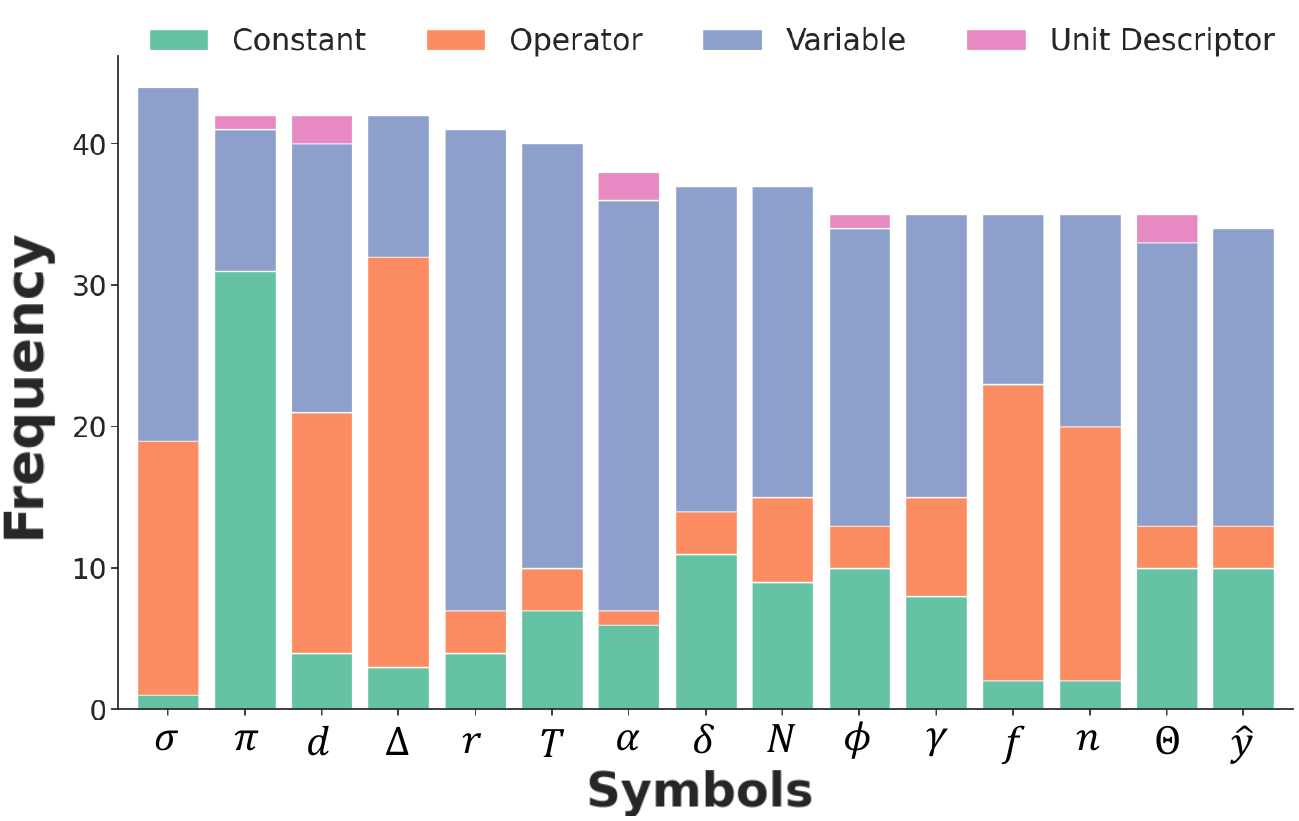}\hfill 
    \caption{Total frequency (appearances) of the top-15 mathematical symbols in 1,000 randomly sampled ArXiv math-rich documents. Each math symbol contemporary belongs to multiple main attribute categories depending on its related context and mathematical expression.
    This illustrates the \textbf{contextual polymorphism} of a single math symbol.}
    \label{fig: frequency}
\end{figure}

Large language models (LLMs) have demonstrated exceptional reasoning abilities across numerous fields \citep{huang2022towards, hadi2023survey, hadi2024large, li2024personal, zheng2024heterogeneous, he2024llm}. With the increasing shift towards applying LLMs to complex tasks \citep{brown2020language, kojima2022large, sui2023tap4llm, zou2024promptintern, he2024cocost}, the need for supplementary data beyond the general pre-trained datasets has become increasingly important.
Among these, mathematical reasoning tasks \cite{english2013mathematical, ilany2010language} have recently drawn the attention of several researchers \cite{imani2023mathprompter, ahn2024large, zhang2024mathverse, lu2023mathvista}. 
In particular, Part-of-Math Tagging \cite{youssef2017part}, the mathematical analog to part-of-speech tagging \cite{schmid1994part}, where mathematical tokens are classified according to a given taxonomy of attributes, continues to gain interest with the integration of LLMs math reasoning.

However, despite this growing interest, Part-of-Math Tagging currently still lacks the foundational datasets that are crucial for supporting advanced NLP tasks \cite{youssef2017part, shan2021towards, shan2024using}.
In addition, integrating mathematical language into NLP models remains a substantial challenge \cite{alshamari2020astudy, meadows2022survey}, especially in the realm of document parsing \cite{dridan2013document, lam2008xml, zhang2019gap}. 
Traditional semantic parsing methods such as  LaTeXML \cite{miller2011latexml} often fall short when applied to math-rich documents, where precision and structured syntax are paramount \citep{hamel2022evaluation, paster2023openwebmath, wang2023scibench}. These methods struggle to accurately perform pattern matching between abstract mathematical symbols and their corresponding XML tag notations. 

Similarly, recent advanced LLMs, such as ChatGPT \citep{liu2023summary}, also face difficulties in understanding and reasoning with abstract mathematical symbols due to their contextual polymorphism \cite{ditchfield1994contextual,fiore2013multiversal,murase2023contextual}, as demonstrated in Figure \ref{fig: frequency}.
As an intuitive example, in the linear equation:
\begin{math}
    y = mx + p
\end{math},
$y$ is categorized as a variable. Whereas in the cross-entropy loss function:
\begin{math}
\mathcal{L}(x, y) = - \sum_{i=1}^{N} y_i \log(x_i)
\end{math},
the symbol $y$ represents the fixed target labels, which is considered a constant for a given dataset. 
Without the corresponding contextual information of a mathematical symbol, LLMs are unable to distinguish between different attributes of the symbol and cannot effectively process related mathematical reasoning tasks.
Thus, tagging math symbols within domain-specific contexts is essential for language models.

\begin{figure*}[t]
\centering
\includegraphics[width= 0.85\textwidth]{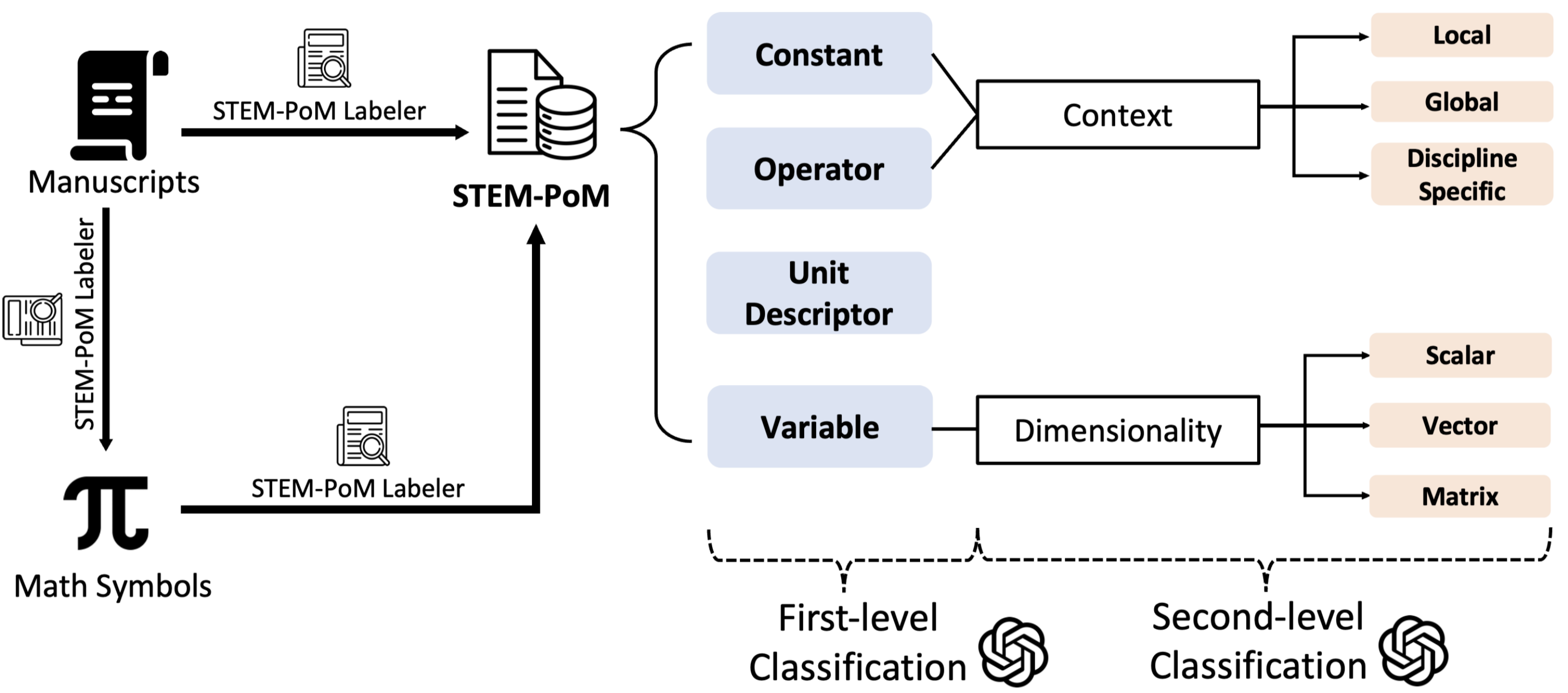}
\caption{The overall pipeline for constructing the \ours dataset. We extract math symbols with corresponding text information to formulate the dataset. Each math symbol is initially classified into one of four primary categories based on its definition. Then, the symbol is further categorized into secondary categories by the context in which it appears or by the symbol's dimensionality. An LLM is evaluated via the first-level and second-level classification tasks.
}
\label{fig: dataset_structure}
\end{figure*}

In this paper, we introduce a novel benchmark dataset, \textbf{\ours}, designed to evaluate the reasoning capabilities of language models on mathematical symbols across different domains. 
The \ours dataset consists of 2,109 mathematical token instances extracted from a random sampling of 10,000 arXiv manuscripts, which are math-rich documents spanning domains such as Mathematics, Physics, Chemistry, and more.
For each dataset instance, we provide a specific mathematical symbol with the corresponding symbol order in the document, main and sub-level attributes, and the related text information from the original ArXiv paper.
Each mathematical symbol in the dataset is classified according to two levels of attributes \cite{wiki_math_symbols}. The first-level attribute categorizes the symbol as variable, constant, operator, or unit descriptor. The second-level attribute further classifies the symbol into one of six types based on its first-level category: scalar, vector, matrix, local, global, or discipline-specific. Figure \ref{fig: dataset_structure} illustrates the category distribution of the dataset. 
To further enrich the \ours dataset with additional arXiv manuscripts and other math-rich document resources, we also design the \textbf{STEM-PoM Labeler}, a feasible toolkit to assist dataset generation by automatically searching, extracting, and recording hand-labeled mathematical symbols with their corresponding context from STEM articles.

We conduct extensive experiments on the \ours dataset to assess the mathematical reasoning abilities of a large amount of open- and closed-source vanilla language models.
Our experimental results indicate that existing language models generally struggle with understanding mathematical symbols, even when provided with relevant context. Extending the context length to a certain degree and fine-tuning on symbol-related material can enhance their ability to classify mathematical symbols. 
We further examine the relationship between the model's mathematical symbol classification and mathematical reasoning capabilities by evaluating its performance on challenging downstream tasks, such as OlympiadBench \cite{he2024olympiadbench}, after enhancing its classification accuracy through fine-tuning on \ours. Last, we provide detailed dataset analysis and case studies to investigate the influence factors such as context length on the ability of language models to understand mathematical symbols. 
In summary, our \textbf{main contributions} are: 
\begin{itemize}[noitemsep, left=0pt]
    \item We propose a novel task combining Part-of-Math Tagging with document parsing to assess LLMs' mathematical symbol classification abilities.

    \item We introduce \ours, a benchmark dataset comprised of over 2,000 annotated instances extracted from math-rich documents spanning fields including Mathematics, Physics, and Chemistry. Each math symbol in the dataset is meticulously labeled with multiple hierarchical attributes, reflecting its role and context within the original document.

    \item We evaluate LLMs' mathematical reasoning on \ours, revealing performance variations across different models and highlighting the role of math symbol understanding in enhancing LLMs' mathematical reasoning abilities.
\end{itemize}

\section{Backgrounds}
\label{appendix: background}

\noindent\textbf{Part-of-Math (PoM) Tagging.} The part-of-math tagging task draws inspiration from similar tagging tasks such as part-of-speech tagging \cite{schmid1994part}. In the PoM context, the goal is to label individual mathematical tokens or expressions in math formulas with their corresponding mathematical roles. 
\citet{youssef2017part} collected mathematical content, such as formula representation and tagging for specific mathematical formula translations and verifications, including converting formulae into semantic LaTeX or testing with tools like CAS (Computer Algebra Systems). However, this focus on structured and narrow formula translations does not align with the broader, more diverse text-based tasks required to assess NLP models, due to the lack of scalability features in the collected math symbols. 
\citet{shan2021towards,shan2024using} recently evaluated the potential of leveraging LLMs for automated annotation and Part-of-Math tagging of math symbols conducted on the Digital Library of Mathematical Functions (DLMF) \cite{lozier2003nist}. Since the source of math symbols is only one manuscript, the mathematical tokens collected only have a single classification type and are self-consistent. In contrast, our dataset incorporates the inherent messiness of published literature across several STEM subjects, where these domain-specific math symbols can have multiple classifications or meanings depending on the discipline and related contextual information.

\noindent\textbf{Math Symbol Annotation.} In the context of mathematical symbol annotations, a common approach is to simply extract definitions for mathematical tokens, transforming part-of-math tagging into a name-entity-recognition (NER) task \cite{asakura2021miogatto, shan2024using}. While NER does evaluate an AI model’s ability to process text, it is fundamentally different from part-of-math tagging, which requires the model to understand and classify mathematical tokens based on a specific taxonomy. Unlike NER, part-of-math tagging tasks defined in our dataset demand deeper comprehension of the mathematical symbol.

\noindent\textbf{Large Language Models.}
Pre-trained large language models (LLMs) have become a cornerstone in modern NLP \cite{rosenfeld2000two, zhao2023survey}. 
Early approaches were based on N-gram models, but with the advent of distributed word embeddings \cite{bengio2000neural, mikolov2013efficient}, neural language models gained prominence. The scalability and performance improvements introduced by these models and the availability of vast textual data have enabled the unsupervised pre-training of LLMs. These models, often referred to as foundation models \cite{radford2019language,kojima2022large}, can then be fine-tuned on smaller, task-specific datasets to adapt them for various downstream applications. For \ours, we apply one traditional sequence-based NLP model, LSTM \cite{graves2012long}, and several most updated LLMs for our dataset evaluation.

\begin{figure*}[t]
\begin{minipage}{0.45\textwidth}
    \centering
    \resizebox{0.8\textwidth}{!}{ 
    \begin{tabular}{lc}
        \toprule
        \textbf{Statistic} & \textbf{Number} \\ \midrule
        \textbf{Total Symbols} & \textbf{2,109} \\
        \midrule
        \textbf{Unit Descriptor} & \textbf{129} \\
        \textbf{Constant} & \textbf{384} \\
        \quad- Local & 171\\
        \quad- Global & 121\\
        \quad- Discipline Specific & 92\\
        \textbf{Operator} & \textbf{363} \\
        \quad- Local & 181\\
        \quad- Global & 105\\
        \quad- Discipline Specific & 77\\
        \textbf{Variable} & \textbf{1,233} \\
        \quad- Scalar & 601\\
        \quad- Vector & 599\\
        \quad- Matrix & 33\\
        \midrule
        Avg symbols per article & 4.7 \\
        Avg tokens per sentence & 31.8 \\
        Avg tokens per math symbol & 1.07\\
        \bottomrule
    \end{tabular}
    }
    \captionof{table}{\ours Dataset Statistics}
    \label{tab: dataset_statistics}
\end{minipage}%
\hfill
\begin{minipage}{0.55\textwidth}
    \centering
    \includegraphics[width=\textwidth]{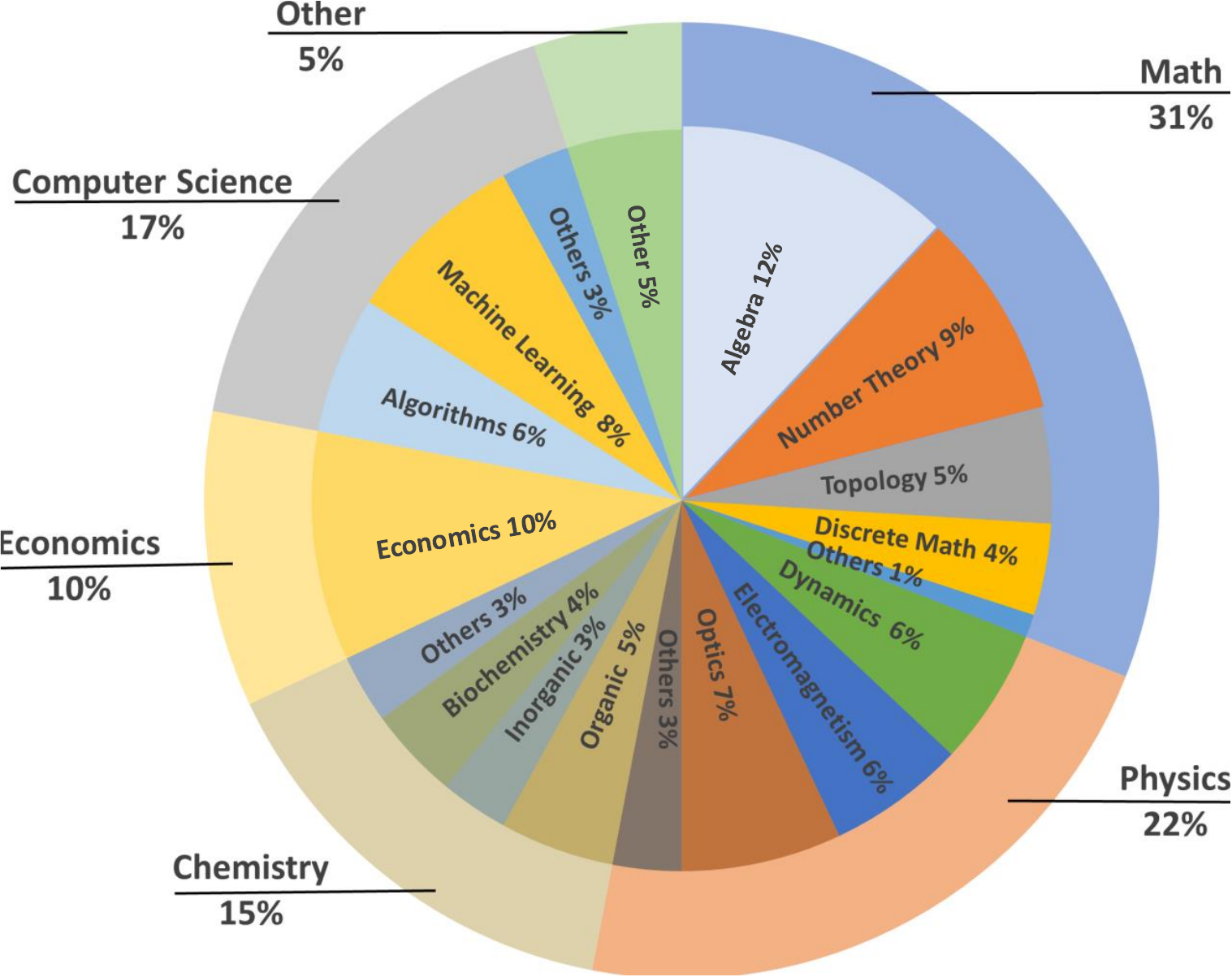} %
    \captionof{figure}{Discipline Distribution from Source ArXiv} 
    \label{fig: arxiv_distribution}
\end{minipage}
\end{figure*}

\begin{table*}[t]
  \centering
  \resizebox{\textwidth}{!}{
    \begin{tabular}{cccccc}
    \toprule
    \textbf{File Name} & \textbf{Symbol Order} & \textbf{Symbol} & \textbf{Main Attribute} & \textbf{Sub Attribute} & \textbf{Related Contents} \\
    \midrule
    9509/adap-org9509001.html & 0 & $f$ & Constant & Global & ...1/$f$ noise was discovered... \\
    9509/adap-org9509001.html & 1 & $\Delta$ & Operator & Global &  ...can be quantified by studying the displacement $\Delta X$ \\
    9509/adap-org9509001.html & 2 & $X$ & Unit Descriptor & - & ...can be quantified by studying the displacement $\Delta X$ \\
    9509/adap-org9509001.html & 3 & $t$ & Variable & Scalar & ..after t steps, we can... \\
    ...&... &... &... &... & ...\\
    \bottomrule
    \end{tabular}
}
\caption{\ours Dataset Structure}
\label{tab: dataset structure}

\end{table*}

\section{\ours Dataset}
In this section, we introduce our constructed benchmark dataset, \ours. First, we describe the source data used for extracting mathematical symbols and text information in Section \ref{sec: source_dataset}. Next, we outline the data annotation process in Section \ref{sec: data_annotation} and present the dataset statistics in Section \ref{sec: dataset_statistics}. Finally, we provide details on our dataset labeling tool STEM-PoM labeler in Section \ref{sec: PoM_Tagger}.
\label{sec:dataset}

\subsection{Source Dataset}
\label{sec: source_dataset}
The first crucial step in constructing the dataset is selecting high-quality mathematical symbols. For \ours, we primarily collect these symbols from two sources: 
1. \textit{Public math-symbol datasets}, where we directly utilize candidate mathematical symbols from the mathematical token definition extraction benchmark, MTDE \citep{hamel2022evaluation}.
2. \textit{Raw ArXiv papers} \cite{clement2019use}, where we identify and extract mathematical symbols with corresponding context including math equations, math symbol definitions, and related text sentences from the dataset. The detailed description of each raw dataset is provided below:

\paragraph{MTDE} \cite{hamel2022evaluation} contains approximately 10,000 entries of mathematical symbol names along with their defined contexts. Each entry includes a 'short' definition and a 'long' definition. A short definition is a single-word definition, while a long definition consists of one or more words. The data was collected through random sampling from mathematical and scientific arXiv preprint manuscripts, covering a broad range of disciplines such as Physics, Computer Science, and Biology. Note that all math symbols in the MTDE dataset are pre-filtered and well-defined by the authors, ensuring our annotation process minimizes the risk of misinformation or repetition issues.
In our dataset pre-processing, we also make sure the candidate data is generated via a corpus crawler and subsequently pruned and cleaned manually.

\paragraph{ArXiv Paper Dataset} \cite{kwarc_arXMLiv} contains 1.7 million arXiv articles, spanning a wide range of disciplines, including Mathematics, Physics, Chemistry, Economics, and Computer Science. We will provide a detailed explanation of how \ours is constructed from the raw source dataset in the next subsection.

\subsection{Dataset Construction} 
\label{sec: data_annotation}

To construct our dataset, we randomly sampled 10,000 articles from the ArXiv Paper Dataset across various domains. We manually verified that each manuscript is math-rich, containing numerous mathematical expressions and symbols. Using pre-filtered mathematical symbols from the MTDE, we engaged human-expert annotators to utilize the STEM-PoM labeler to: (1) align each mathematical symbol with its source paper, (2) identify relevant contextual information for each symbol, and (3) address disambiguation and aliasing by annotating all possible meanings for symbols with multiple interpretations and specifying the context in which each meaning applies. Through the rigorous selection process, 453 of the 10,000 articles are identified as the final source articles with high-quality contextual matches for the symbols. Each article contains an average of 4.7 annotated mathematical symbols (2109 in total).

After obtaining the mathematical symbols, we categorize each symbol into different attributes and assign surrounding context information to construct the \ours dataset. Specifically, we first extract the file name and symbol order for each mathematical symbol. Then, for each symbol, we extract the contexts in which the symbol appears, using several predefined lengths. Following this, we manually classify each symbol into four main attribute (first-level) categories: Variable, Constant, Operator, and Unit Descriptor. 
For the Variable, Constant, and Operator, we further categorize them into sub-attribute (second-level) categories. A variable is classified as a Vector, Scalar, or Matrix, while a Constant or Operator is categorized as Local, Global, or Discipline-Specific. Table \ref{tab: dataset structure} outlines the overall dataset structure. We manually examine each entry of the dataset thoroughly to ensure its robustness and correctness. 
The overall dataset pipeline is demonstrated in Figure \ref{fig: dataset_structure}.
We also provide a detailed explanation of the dataset structure in Appendix \ref{appedix: dataset_definition} and the definitions of each level's attributes in Appendix \ref{appendix: attribute_definition}. 

\paragraph{Dataset Statistics}
\label{sec: dataset_statistics}
We summarize the key statistics of our dataset in this section. Table \ref{tab: dataset_statistics} presents the categorical statistics, including the math symbols along with their first- and second-level attributes. The distribution of Variables, Constants, Operators, and Unit Descriptors is 58.5\%, 18.2\%, 17.2\%, and 6.1\%, respectively. In addition, Figure \ref{fig: arxiv_distribution} illustrates the discipline distribution of the source arXiv papers. Our dataset covers mathematical symbols from various fields, including Mathematics, Physics, Chemistry, Economics, Computer Science, etc.

\subsection{Dataset Quality Control} 
During the dataset construction, we manually evaluate the quality and applicability of the annotated data. Specifically, we process the evaluation process through both consistency checks and Inter-annotator agreements.

\paragraph{Consistency Checks.} To ensure reliability, we perform the following consistency checks: each article is assigned to domain-specific experts for labeling, and we provide precise definitions and examples for each label and category to guide the process. After labeling, we randomly select a subset of the labeled data for consistency evaluation. Additionally, we analyze data points with inconsistent labels, facilitate discussions among annotators to resolve ambiguities and reach a consensus, and refine the annotation guidelines to address any common sources of confusion.

\paragraph{Inter-annotator Agreements.} Regarding the IAA in our human annotation process, we engage 33 domain experts as our annotators to annotate domain-specific articles. After the initial labeling process, the rest annotators (5 Annotator/Discipline on average) from the same domain reviewed the annotated data by switching their labeling components. For the Inter-annotator Agreement, we measure with Cohen's Kappa value and ensure the value ranges from 0.81 to 1 (0.903 on average across all labeled data), regarding the dataset's mathematical symbols and their corresponding attributes. To maintain accuracy, annotators have access to the original papers through the STEM-PoM labeler and annotate the attributes of the data based on the content of the original papers. This thorough review process reinforces the reliability of our annotated dataset.

\subsection{STEM-PoM Labeler}
\label{sec: PoM_Tagger}
We developed a toolkit named STEM-PoM Labeler to assist human experts in retrieving math symbols along with their corresponding article contexts, annotating the dataset, and conducting inter-annotator consistency checks to ensure agreement in dataset labeling. For a detailed description of the toolkit's design, please refer to Appendix \ref{appendix: UI_Design}.

\section{Experiments}
\label{sec:results}

\begin{table*}[t]
\centering 
\resizebox{\linewidth}{!}{
\begin{tabular}{llccccc}
\toprule
\textbf{Models} & \textbf{Context Length} & \textbf{Overall} & \textbf{Variable} & \textbf{Constant} & \textbf{Operator} & \textbf{Unit Descriptor} \\
\midrule
\multirow{3}{*}{LSTM \cite{graves2013generating}} 
& One Sentence & 18.7\% & 24.5\% & 13.2\% & 10.3\% & 27.1\% \\
& Ten Sentences & 22.6\% & 28.1\% & 16.8\% & 15.5\% & 30.2\% \\
& Full Manuscript & - & - & - & - & - \\
\midrule
\multirow{3}{*}{Llama2-13B \cite{touvron2023llama}} 
& One Sentence & 36.8\% & 24.1\% & 39.3\% & 41.4\% & 42.7\% \\
& Ten Sentences & 42.7\% & 35.6\% & 39.8\% & 46.9\% & 48.5\% \\
& Full Manuscript & 45.9\% & 38.2\% & 42.8\% & 50.1\% & 52.4\% \\
\midrule
\multirow{3}{*}{Mistral-8x7B \cite{jiang2024mixtral}} 
& One Sentence & 47.3\% & 38.5\% & 41.7\% & 52.9\% & 56.2\% \\
& Ten Sentences & 49.8\% & 41.8\% & 45.9\% & 58.6\% & 56.7\% \\
& Full Manuscript & 53.6\% & 45.7\% & 48.9\% & 61.4\% & 58.2\% \\
\midrule
\multirow{3}{*}{Llama3.1-70B \cite{dubey2024llama}} 
& One Sentence & 48.9\% & 41.3\% & 44.6\% & 48.5\% & 61.5\% \\
& Ten Sentences & 53.0\% & 44.8\% & 48.8\% & 54.7\% & 63.7\% \\
& Full Manuscript & 51.7\% & 42.7\% & 43.2\% & 55.2\% & 65.8\% \\
\midrule
\multirow{3}{*}{Claude3.5-Sonnet \cite{anthropic2024}} 
& One Sentence & 63.7\% & 58.6\% & 62.5\% & 65.7\% & 67.8\% \\
& Ten Sentences & 65.9\% & 61.3\% & 64.3\% & 67.9\% & 70.2\% \\
& Full Manuscript & 66.7\% & 62.9\% & 65.8\% & 68.6\% & 69.3\% \\
\midrule
\multirow{3}{*}{GPT-3.5-turbo \cite{achiam2023gpt}} 
& One Sentence & 56.8\% & 51.5\% & 53.5\% & 59.4\% & 62.4\% \\
& Ten Sentences & 58.7\% & 54.5\% & 53.6\% & 61.3\% & 65.1\% \\
& Full Manuscript & 60.6\% & 57.2\% & 56.6\% & 63.2\% & 65.2\% \\
\midrule
\multirow{3}{*}{GPT-4o \cite{hurst2024gpt}} 
& One Sentence & 64.9\% & 60.5\% & 64.2\% & 64.9\% & 70.1\% \\
& Ten Sentences & 67.4\% & 63.7\% & 66.1\% & 66.4\% & 73.5\% \\
& Full Manuscript & 68.5\% & 64.2\% & 67.8\% & 68.1\% & 73.8\% \\
\bottomrule
\end{tabular}
}
\caption{First-level classification accuracy with various context lengths. Here. \textbf{One sentence/Ten Sentences/Full Manuscript} refers to the context size of completed sentences as contextual information for each math symbol.}

\label{tab:first_level}
\end{table*}

\begin{table*}[!t]
\centering 
\resizebox{\linewidth}{!}{
\begin{tabular}{lccccccccc}
\toprule
\multirow{2}{*}{\textbf{Models}} & \multicolumn{3}{c}{\textbf{Variable}} & \multicolumn{3}{c}{\textbf{Constant}} & \multicolumn{3}{c}{\textbf{Operator}} \\ 
\cmidrule(lr){2-4} \cmidrule(lr){5-7} \cmidrule(lr){8-10}
& \textbf{Scalar} & \textbf{Vector} & \textbf{Matrix} & \textbf{Local} & \textbf{DS} & \textbf{Global} & \textbf{Local} & \textbf{DS} & \textbf{Global} \\ \midrule
LSTM \cite{graves2013generating} & 13.8\%           & 15.1\%          & 17.2\%          & 19.2\%         & 17.8\%                     & 22.2\%         & 16.6\%         & 11.3\%                     & 14.6\%         \\ \midrule
Llama2-13B \cite{touvron2023llama} & 27.3\%& 24.4\%& 21.8\% & 33.6\%& 31.5\%& 33.6\%& 32.4\%& 28.3\%& 32.7\%\\ \midrule
Mistral-8x7B \cite{jiang2024mixtral} & 36.9\%    & 35.8\%          & 21.6\%          & 34.8\%         & 31.2\%                     & 37.8\%         & 36.4\%         & 34.8\%                     & 35.7\%         \\ \midrule
Llama3.1-70B \cite{dubey2024llama} & 38.2\% & 34.1\% & 26.7\% & 37.6\% & 35.2\% & 36.1\% & 39.1\% & 32.3\% & 40.2\% \\ \midrule

Claude3.5-Sonnet \cite{anthropic2024} & 53.2\% & 49.7\% & 55.8\% & 55.9\% & 53.1\% & 49.6\% & 56.3\% & 52.2\% & 55.9\%\\ \midrule
GPT-3.5-turbo \cite{achiam2023gpt} & 44.5\%         & 45.8\%          & 48.3\%          & 48.5\%         & 42.9\%                     & 44.3\%         & 48.4\%         & 43.5\%                     & 49.7\%         \\  \midrule
GPT-4o \cite{hurst2024gpt} & 54.6\% & 51.3\% & 58.6\% & 58.4\% & 54.1\% & 56.2\% & 60.5\% & 57.3\% & 58.5\% \\ 

\bottomrule
\end{tabular}
}
\caption{
Second-level classification accuracy with full manuscript input (Ten-sentence input for LSTM). We abbreviate "Discipline Specific" as "DS".
}
\label{tab:second_level}
\vspace{-10pt}
\end{table*}

\subsection{Setups}
\paragraph{Models.} 
To thoroughly evaluate our dataset across models with varying parameter sizes, we utilize the following models: LSTM \cite{graves2013generating}, Mixtral-8x7B-v0.1 \cite{jiang2024mixtral}, Llama2-13B \cite{touvron2023llama}, Llama3.1-70B \cite{dubey2024llama}, Claude-3.5-Sonnet \cite{anthropic2024}, GPT-3.5-Turbo-0125\footnote{\url{https://platform.openai.com/docs/models/gpt-3-5-turbo}} \cite{achiam2023gpt}, and GPT-4o-2024-08-06\footnote{\url{https://platform.openai.com/docs/models/gpt-4o}} \cite{hurst2024gpt}.

\paragraph{Evaluation Metrics.} We apply the \textit{Precision Accuracy} as our metric for the mathematical symbol classification task, the metric can be formulated as:
\begin{equation*}
    \small 
    \textit{Precision Accuracy} = \frac{\textit{Number of correct predictions}}{\textit{Total number of samples}}
\end{equation*}

\paragraph{Training \& Inference Details.} 
We evaluate several models under both pre-training and fine-tuning settings. Specifically, we train an LSTM model with varying layers and apply the LoRA method, a parameter-efficient fine-tuning (PEFT) technique, to several updated LLMs. We also evaluate additional LLMs under the in-context learning setting. The training and model parameter details are provided in Appendix \ref{appendix: training_details}.
\begin{table*}[t]
\centering 
\resizebox{\linewidth}{!}{
\begin{tabular}{llccccc}
\toprule
\textbf{Models (fine-tuned)} & \textbf{Context Length} & \textbf{Overall} & \textbf{Variable} & \textbf{Constant} & \textbf{Operator} & \textbf{Unit Descriptor} \\
\midrule
Llama-2-13B \cite{touvron2023llama}
& Ten Sentences & 45.5\% & 38.4\% & 41.7\% & 50.6\% & 51.3\% \\
\midrule
Mixtral-8x7B \cite{jiang2024mixtral}
& Ten Sentences & 55.7\% & 46.3\% & 51.9\% & 64.2\% & 60.3\% \\
\midrule
Llama3.1-70B \cite{dubey2024llama}
& Ten Sentences & 62.4\% & 56.6\% & 54.5\% & 70.8\% & 67.5\% \\
\midrule
GPT-3.5-turbo \cite{achiam2023gpt}
& Ten Sentences & 66.9\% & 65.4\% & 66.6\% & 71.3\% & 64.5\% \\
\midrule
GPT-4o \cite{hurst2024gpt}
& Ten Sentences & 70.3\% & 71.3\% & 74.1\% & 75.2\% & 68.1\% \\
\bottomrule
\end{tabular}
}
\caption{First-level classification accuracy with various context lengths. Here. "One sentence/Ten Sentences/Full Manuscript" refers to the context size of completed sentences used as contextual information for testing each mathematical symbol.}

\label{tab:first_level_finetune}
\vspace{-5pt}
\end{table*}

\vspace{-7pt}
\subsection{First-Level Classification Results}
Table \ref{tab:first_level} summarizes the accuracy results of various models across different context lengths. To extract context information, we implement a surrounding window provided for domain human expert annotators to meticulously select the most relevant completed sentences surrounding the math symbol. Through the careful context selection process, we ensure that the information included in the prompts is accurate and contextually relevant. The result shows that the small-parameter-size language model such as the LSTM struggles with lower accuracy, achieving between 18.7\% and 22.6\%. 
In contrast, larger models, such as Claude3.5-Sonnet and GPT-4o, show marked improvements as context length increases, with accuracy consistently above 63.7\% and up to 73.8\%.
We also found that the performance gap between models remains consistent as context length increases. To demonstrate, GPT-4o outperforms Llama3.1-70B by 16.0\%, 14.4\%, and 16.8\% for context lengths of one sentence, ten sentences, and the full manuscript, respectively. This consistent performance gap suggests that larger models with more pre-trained knowledge, such as GPT-4o, exhibit superior scalability with longer contexts. 

Another notable observation is that the overall performance gain from increasing context length is more pronounced in smaller models, such as Llama2-13B and Mistral-8x7B, which have less pre-trained knowledge. These models benefit more from extended context as they rely on additional information to compensate for their limited pre-training. Larger models like GPT-4o and Claude3.5-Sonnet, which come with extensive pre-trained knowledge, show relatively smaller performance gains as context length increases.

\subsection{Second-Level Classification Results.}
Table \ref{tab:second_level} shows second-level classification accuracy with full manuscript input. In this experiment setting, we assume that the model got the first-level classification correct.
By horizontally comparing the same model performance on different sub-attribute classifications, we find that the attribute Constants are generally easier to classify compared to Variables and Operators across all sizes of models, as seen by the overall higher accuracy in Constant-related tasks. However, Matrix and DS classification continue to present challenges, even for the largest models, indicating that certain structures, as well as content types within manuscripts, remain difficult to categorize accurately at the sub-attribute level.

\label{sec: overall}
\noindent\textbf{Overall,} performance across all models on both first-level and second-level classification tasks shows a clear trend of improvement with increasing context length, highlighting the importance of context for accurately classifying mathematical symbols.
Additionally, both small and large-size language models show a relatively higher accuracy in identifying Unit Descriptors and Operators compared to Variables and Constants, indicating that symbols with more distinct contextual or syntactical patterns are easier for models to classify. Through the above results, we aim to gain insights into the extent to which different category attributes of mathematical symbols influence LLMs' understanding of math-rich documents by correctly classifying the symbols in real-world scenarios. 

\subsection{Fine-tuning on \ours}
\label{sec: finetune}
We fine-tune four different sizes of LLMs on \ours, as detailed in Table \ref{tab:first_level_finetune}. Comparing these results with those in Table \ref{tab:first_level}, we observe that additional training improves model classification accuracy for mathematical symbols when provided with more context information. However, continuously adding contextual knowledge to the training samples does not always lead to performance gains. We analyze this relationship further in Section \ref{sec: ablation_study}.

\subsection{Downstream Math Reasoning}
After evaluating the performance of \ours across different LLMs, we further investigate its effectiveness in enhancing LLMs' math reasoning capabilities. Specifically, we aim to address the following questions:

\begin{itemize} [noitemsep, left=0pt]
    \item \textbf{Q1:} What is the relationship between an LLM’s ability to classify mathematical tokens and its real-world mathematical reasoning skills?
    \item \textbf{Q2:} Does improving an LLM’s mathematical token classification capability enhance corresponding math reasoning abilities?
\end{itemize}

\definecolor{darkgreen}{RGB}{0, 128, 0}
\begin{table}[!h]
\centering
\resizebox{\linewidth}{!}{
\begin{tabular}{l|lll|l}
\toprule
Models & GSM8K & MATH & OlympiadBench & \textbf{Avg.}\\ 
\midrule
Llama2-13B & 42.5\% & 29.1\% & 11.5\% & 27.7\% \\
+ LoRA (\ours) & 44.6\% (\textcolor{darkgreen}{\(\uparrow 2.1\)}) & 31.3\% (\textcolor{darkgreen}{\(\uparrow 2.2\)}) & 13.4\% (\textcolor{darkgreen}{\(\uparrow 1.9\)}) & 29.8\% (\textcolor{darkgreen}{\(\uparrow 2.1\)}) \\
\midrule
Mixtral-8x7B & 72.4\% & 32.6\% & 13.7\% & 39.6\% \\
+ LoRA (\ours) & 74.1\% (\textcolor{darkgreen}{\(\uparrow 1.7\)}) & 34.1\% (\textcolor{darkgreen}{\(\uparrow 1.5\)}) & 16.4\% (\textcolor{darkgreen}{\(\uparrow 2.7\)}) & 41.5\% (\textcolor{darkgreen}{\(\uparrow 1.9\)}) \\
\midrule
Llama3.1-70B & 91.6\% & 47.1\% & 26.4\% & 55.0\% \\
+ LoRA (\ours) & 93.2\% (\textcolor{darkgreen}{\(\uparrow 1.6\)}) & 48.8\% (\textcolor{darkgreen}{\(\uparrow 1.7\)}) & 28.2\% (\textcolor{darkgreen}{\(\uparrow 1.8\)}) & 56.7\% (\textcolor{darkgreen}{\(\uparrow 1.7\)}) \\
\midrule
GPT-4o & 94.3\% & 88.7\% & 39.6\% & 74.2\% \\
+ LoRA (\ours) & 95.2\% (\textcolor{darkgreen}{\(\uparrow 0.9\)}) & 88.9\% (\textcolor{darkgreen}{\(\uparrow 0.2\)}) & 41.2\% (\textcolor{darkgreen}{\(\uparrow 1.6\)}) & 75.1\% (\textcolor{darkgreen}{\(\uparrow 0.9\)}) \\
\bottomrule
\end{tabular} 
}
\caption{Evaluation of LLMs on downstream mathematical reasoning tasks. \textbf{+ LoRA (\ours)} indicates that the corresponding model is first fine-tuned using LoRA on \ours before being evaluated on downstream mathematical reasoning tasks. The improvements from LoRA fine-tuning on \ours are highlighted in \textcolor{darkgreen}{darkgreen} (\(\uparrow x\)).}
\label{tab:math_reason_res}
\end{table}

\noindent To investigate these questions, we select challenging mathematical problems from GSM8K \cite{cobbe2021training}, MATH \cite{hendrycks2021measuring}, and OlympiadBench \cite{he2024olympiadbench} as our downstream tasks. We evaluate the model's performance before and after LoRA fine-tuning on \ours. The evaluation is conducted using greedy decoding without tool integration. The primary metric reported is pass@1 accuracy with 3-shots CoT prompting. Full implementation details are shown in Appendix \ref{appendix: downstream}.

Table \ref{tab:math_reason_res} presents the experimental results on three mathematical reasoning datasets. A comparison across different models reveals that LLMs demonstrate superior performance on STEP-PoM and also generally achieve higher accuracy in mathematical reasoning tasks. This suggests that \textbf{higher math token classification accuracy generally correlates with better mathematical reasoning performance.} Additionally, all models exhibit performance gains after being fine-tuned on \ours. This suggests that \textbf{enhancing LLMs’ understanding of mathematical symbols contributes to improved reasoning abilities in mathematical problem-solving.} The experimental results align with the objective of our constructed dataset, which aims to fundamentally enhance LLMs' mathematical reasoning through Part-of-Math Tagging.

\subsection{Analysis on \ours}
\label{sec: ablation_study}
\paragraph{Model Performance vs Model Size.} 
\begin{table}[!ht]
\centering
\resizebox{\linewidth}{!}{
\begin{tabular}{ccccc}
\toprule
Model size(layers) & Variable & Constant & Operator & Unit Descriptor \\ \midrule
128      & 24.5\%            & 13.2\%           & 10.3\% & 27.1\%                   \\ 
256    & 28.7\%            & 17.9\%            & 15.7\%           & 32.5\%                   \\ 
512       & 34.2\%            & 23.2\%            & 24.9\%            & 40.0\%                     \\ 
1024        & 46.5\%           & 35.9\%           & 44.2\%           & 51.3\%
     \\ \bottomrule
\end{tabular}
}
\caption{LSTM first-level classification accuracy based on different model sizes}
\label{tab: model_sizes}
\end{table}

Table \ref{tab: model_sizes} presents the classification accuracy of an LSTM model for first-level classification across different model sizes, ranging from 128 to 1024 layers. Note that we set the input context length to be one sentence. The results show a clear positive correlation between the model size and classification accuracy across all four categories. For the smallest model (128 layers), the accuracy ranges from 10.3\% for the Operator class to 27.1\% for the Unit Descriptor class. As the model size increases, the performance improves notably, with the largest model (1024 layers) achieving a relatively high-performance gain in accuracy, ranging from 35.9\% for the Constant class to 51.3\% for the Unit Descriptor class. The most substantial improvements are observed in the Operator category, where accuracy increases from 10.3\% for 128 layers to 44.2\% for 1024 layers. These results suggest that larger model sizes are more effective in capturing complex patterns. 
We further analyze the impact of input sequence length on model performance and compare the effectiveness of fine-tuning versus in-context learning on \ours. Please refer to \ref{appendix:full_analysis} for full analysis.

\paragraph{Case Study and Further Discussion.}
Please refer to Appendix \ref{appendix: case_study} and \ref{appendix: discussion} for the additional case study and discussions on \ours.

\section{Conclusion}
In this paper, we introduce \ours, a comprehensive benchmark for evaluating language models' mathematical reasoning abilities to classify math symbols from scientific texts. The dataset includes over 2,000 math instances sourced from ArXiv papers. Extensive experiments show that the best-performing model, achieves only 73.8\% and 60.5\% for first and second-level classification accuracy, highlighting the challenge of extracting and categorizing math symbols from large text corpora.

\section{Limitations}
While we introduce a new benchmark dataset to evaluate LLMs' reasoning abilities with math symbols, several challenges identified in our experimental evaluations warrant further investigation. First, as discussed in Section \ref{sec: overall}, even the most advanced LLMs struggle to accurately recognize complex math symbol attributes, such as matrices and discipline-specific constants or operators. To address this limitation, additional prompting or encoding methods, such as \citep{fu2023edsl}, need to be explored. Second, as evidenced by Table \ref{tab:first_level} and \ref{tab:second_level}, when computational resources are constrained, leveraging smaller language models to process extensive mathematical context while maintaining performance levels comparable to larger models like GPT-4o presents a significant challenge. Through our benchmark, \ours, we intend to investigate these directions in future work to enhance LLMs' mathematical reasoning capabilities within the scope of Part-of-Math Tagging.

\section{Ethics Statement}
The dataset proposed in this paper is derived from publicly available datasets that were released under licenses permitting reuse for research purposes. In constructing our dataset, we have taken care to ensure that all data sources comply with applicable legal and ethical standards, including privacy and data protection regulations. We have only used datasets that are free from personally identifiable information (PII) and sensitive attributes and have not modified or added any data that could compromise the anonymity or privacy of individuals. Furthermore, we encourage responsible use of the dataset and have made it available solely for academic and research purposes, with the expectation that future users will adhere to similar ethical guidelines.

\bibliography{main}

\newpage
\appendix
\appendix
\section{\ours Dataset Supplementary Materials}

\subsection{Dataset Definitions in Table \ref{tab: dataset structure}}
\label{appedix: dataset_definition}
\textbf{File Name:} This attribute serves as a reference point, indicating the source of the file. Specifically, it denotes the arXiv article from which the dataset extracts its contents.

\noindent
\textbf{Symbol Order:} This component records the sequence in which mathematical symbols appear within the article. By capturing the ordinal position of each symbol, we facilitate a structured analysis of the symbols' progression and their contextual relationships within the document.

\noindent
\textbf{Symbols:} This field encapsulates the mathematical symbols themselves, predominantly consisting of Greek letters, albeit inclusive of additional characters. The precise documentation of these symbols is paramount for the subsequent analytical processes.

\noindent
\textbf{Main and Sub Attributes:} These attributes categorize each mathematical symbol into specific classes, delineating a hierarchical structure within the dataset. This classification scheme is vital for understanding the symbols' roles and relationships within the mathematical discourse.

\noindent
\textbf{Related Contents:} This segment comprises the words or sentences surrounding each symbol, embodying a critical resource for our model training. The contextual information surrounding each symbol is indispensable, as it imbues our models with a deeper understanding of each symbol's application and significance within the mathematical narrative.

\subsection{First-Level and Second-Level Attributes Definition}
\label{appendix: attribute_definition}
\textbf{Constant:} A value that does not change in a mathematical expression.
\textbf{Local Constant:} Constant that is specific to a particular system or model, such as the gravitational constant in a simulation of a specific planetary system.
\textbf{Global Constant:} Constant that is applicable in all contexts, like the speed of light in a vacuum.
\textbf{Discipline-specified Constant:} Constant that applies to particular fields of study, for instance, Planck's constant in quantum mechanics.
\textbf{Operator:} A symbol that operates on one or more operands.
\textbf{Local Operators:} Operator that is applied in a localized or specific context within a discipline, like a self-defined operation in matrix processing.
\textbf{Global Operators:} Operators that is used broadly across different disciplines, like the addition or multiplication operator.
\textbf{Discipline-specified Operators:} Operator that is unique to certain fields, such as the Hamiltonian operator in quantum physics.
\textbf{Variable:} A symbol that represents an unknown or changeable quantity in a mathematical expression.
\textbf{Scalar:} A quantity that has only magnitude, no direction. 
\textbf{Vector:} A quantity that has both magnitude and direction.
\textbf{Matrix:} A rectangular array of numbers or symbols arranged in rows and columns.

\subsection{STEM-PoM Labeler}
\label{appendix: UI_Design}
\begin{figure}[!htp]
\resizebox{!}{\linewidth}{
\centering
\includegraphics[width=\textwidth]{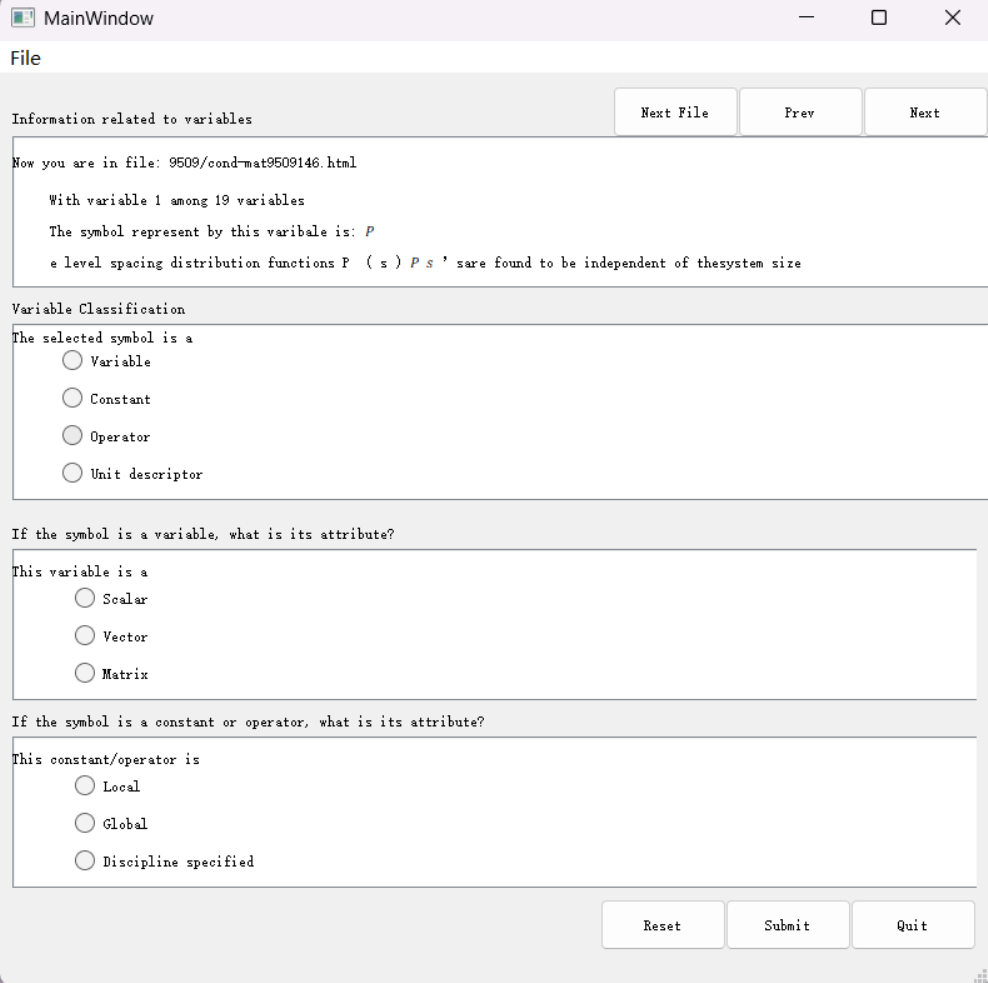}
}
\caption{The UI Design of STEM-PoM Labeler}
\label{fig: ui_design}
\end{figure}

During the dataset construction, a pivotal step involves the meticulous annotation of each mathematical symbol with corresponding tags. This process, inherently labor-intensive and repetitive, necessitates a systematic approach to mitigate the workload and facilitate collaboration among the research team members. To address these challenges, we developed a labeling pipeline designed to streamline the dataset construction process. The UI design is shown in Figure \ref{fig: ui_design}. The functionalities are delineated below:

\paragraph{File Reading.} We initiate the data importing operation progress by importing files from the designated arXiv folder, ensuring a structured and accessible repository of mathematical documents for subsequent processing.

\paragraph{Symbol Identification and Contextualization.} For each file, we enumerate and display essential information: the current file being processed, the total number of symbols within, the sequence number of the current symbol, the graphical representation of the symbol, and the contextual content surrounding the symbol. This feature aids in providing a comprehensive overview and facilitates accurate symbol annotation.

\paragraph{Annotation Interface.} We then present a user-friendly interface offering a set of predefined tagging options for each symbol. Through the designed interface, we easily select the most appropriate tag from these options, standardizing the labeling process and enhancing the consistency of the dataset.

\paragraph{Data Recording.} Upon the selection of a tag for a symbol, We record this association by appending a new line to the dataset, capturing the symbol along with its assigned tag. This systematic data recording ensures the integrity and scalability of the MTCE dataset.

\section{Additional Experiment Setups}
\label{appendix: training_details}
\textbf{Training Details.}
In our experiments, we train an LSTM with varying numbers of layers for the mathematical symbol classification tasks. For LLMs fine-tuning, we apply the common parameter-efficient fine-tuning (PEFT) method, LoRA \cite{hu2021lora}, to evaluate the model precision performance. Specifically, we set the LoRA rank to 32, batch size to 32, weight decay to 0.01, dropout to 0.1, and learning rate to $1e^{-4}$.

\paragraph{Dataset Split.} We divide \ours into 80\%/10\%/10\% for training/validation/testing sets. 

\paragraph{Model Details.} For the LSTM model, we use different layer sizes from \{128, 256, 512, 1024\}. The hidden state size is set to 256, the learning rate is set from \{0.1, 0.01, 0.001\}, the training epoch is 5, and the batch size is 16. We utilize the Adam optimizer \cite{kingma2014adam}. 
For LLMs generation, we set the temperature for all models to 0 (Greedy Decoding), top\_p to 1.0, frequency penalty to 0, and presence penalty to 0.

\section{Downstream Math Reasoning}
\label{appendix: downstream}
\textbf{Evaluation Details.} We use a unified 3-shot CoT prompt template for all models, provided in Figure \ref{fig: math_reasoning_prompt}. Decoding is performed with a temperature of 0 (Greedy Decoding), and we report pass@1 accuracy. Dataset details are outlined below.

\begin{itemize}
\item \textbf{GSM8K} \cite{cobbe2021training} is a mathematical dataset comprising 8.5K high-quality, linguistically diverse grade-school math word problems designed for multi-step reasoning. Solutions involve elementary arithmetic operations and require no concepts beyond early algebra. The test set consists of 1,319 unique problems.

\item \textbf{MATH} \cite{greiner2022math} is a dataset of 12,500 challenging competition-level mathematics problems sourced from contests such as AMC 10, AMC 12, and AIME. Each problem is accompanied by a step-by-step solution, enabling models to learn answer derivations and explanations. The test set comprises 5,000 unique problems.

\item \textbf{OlympiadBench} \cite{he2024olympiadbench} is a bilingual, multimodal scientific benchmark comprising 8,476 Olympiad-level math and physics problems, including those from the Chinese college entrance exam. For our evaluation, we use the open-ended, text-only math competition subset, which consists of 674 problems.

\end{itemize}
\paragraph{Implementation Details.} For LLMs generation, we set the temperature for all models to 0 (Greedy Decoding), top\_p to 1.0, frequency penalty to 0, and presence penalty to 0.

\section{Full Analysis on \ours}
\label{appendix:full_analysis}
\begin{table}[!h]
\centering
\resizebox{\linewidth}{!}{
\begin{tabular}{ccccc}
\toprule
Context Length & Variable & Constant & Operator & Unit Descriptor \\ 
\midrule
One Sentence & 24.5\% & 13.2\% & 10.3\% & 27.1\% \\ 
Five Sentence & 26.3\% & 15.6\% & 14.1\% & 29.2\% \\ 
Ten Sentence & 28.1\% & 16.8\% & 15.5\% & 30.2\% \\ 
\bottomrule
\end{tabular} 
}
\caption{LSTM first-level classification accuracy based on different input context lengths. }
\label{tab: input_length}
\end{table}

\paragraph{Model Performance vs Data Input Lengths.}
Table \ref{tab: input_length} displays the classification accuracy of an LSTM model across varying input context lengths across four categories. A trend of increasing accuracy can be observed as the input length increases. For instance, in the Variable category, the accuracy increases from 24.5\% for one sentence to 28.1\% for ten sentences. Similarly, for the Constant category, accuracy rises from 13.2\% for one sentence to 16.8\% for ten sentences. The Operator category shows a modest increase from 10.3\% to 15.5\% as the input length expands. Finally, for the Unit Descriptor category, accuracy grows from 27.1\% to 30.2\%. These results suggest that longer input data contributes to improved classification accuracy.

\begin{table}[!ht]
\centering 
\resizebox{\linewidth}{!}{
\begin{tabular}{lccccc}
\toprule
\textbf{Context Length} & \textbf{Overall} & \textbf{Variable} & \textbf{Constant} & \textbf{Operator} & \textbf{Unit Descriptor} \\
\midrule
\multicolumn{6}{c}{\textbf{\textit{Vanilla Inference}}}  \\ 
\midrule 
One Sentence & 56.8\% & 51.5\% & 53.5\% & 59.4\% & 62.4\% \\
Ten Sentences & 58.7\% & 54.5\% & 53.6\% & 61.3\% & 65.1\% \\
Full Manuscript & 60.6\% & 57.2\% & 56.6\% & 63.2\% & 65.2\% \\
\midrule
\multicolumn{6}{c}{\textbf{\textit{LoRA Fine-tuned}}}  \\ 
\midrule
One Sentence & 67.4\% & 64.8\% & 67.5\% & 71.4\% & 66.1\% \\
Ten Sentences & 66.9\% & 65.4\% & 66.6\% & 71.3\% & 64.5\% \\
Full Manuscript & 62.2\% & 58.4\% & 62.2\% & 65.1\% & 63.2\% \\
\bottomrule
\end{tabular}
}
\caption{First-level classification with various context lengths on GPT-3.5 and fine-tuned GPT-3.5.}
\label{tab:fine_tune}
\end{table}

\paragraph{Fine-tuning vs In-context learning.} 
Table \ref{tab:fine_tune} shows the comparison result on main attributes between fine-tuned and directly vanilla-referenced GPT3.5. Notably, the fine-tuned GPT-3.5 model achieves an accuracy of 67.4\% in the one-sentence context. However, its performance diminishes as the context length increases, with a noticeable drop to 66.9\% for ten sentences and further down to 62.2\% for full manuscript-length context. 
Such diminishing return for fine-tuned models with longer contexts indicates that fine-tuning amplifies sensitivity to the introduction of noisy or less relevant information when longer contexts are involved. The observation also could point to challenges in the fine-tuning process for long-context LLMs, which require more refined techniques to handle context length effectively.

\section{Case Study on \ours}
\label{appendix: case_study}
To further demonstrate the utility and effectiveness of the STEM-POM dataset, we conducted a case study focusing on error-prone classification scenarios involving mathematical symbols with ambiguous or context-dependent attributes. Our main goal for this case study is to evaluate where and why current state-of-the-art language models fail in complex mathematical reasoning tasks.

\subsection{Setup}

\paragraph{Sub-dataset Selection.}
Building on the main experimental results, we select a subset of mathematical symbols from \ours dataset—particularly those frequently misclassified—to analyze the model's failure modes.

\paragraph{Evaluation Metrics.}
We focused on precision accuracy and the frequency of specific errors across the selected symbols. Additionally, we conducted qualitative error analysis to identify consistent failure cases.

\paragraph{Experimental Contexts.}
Two context lengths were considered for testing:
\begin{itemize}
    \item \textbf{Ten Sentences}: Top Ten sentences selected by human experts surrounding the symbol.
    \item \textbf{Full Manuscript}: Entire input manuscript.
\end{itemize}

\subsection{Results and Analysis}

\paragraph{Error Frequency and Common Failure Cases.}
Table \ref{tab:error_symbols} highlights the top ten symbols that LLMs frequently misclassified. The error rates are calculated as the percentage of incorrect predictions across all contexts. Notably, even the best-performing model, GPT-4o, exhibited a significant error rate for symbols with overlapping roles, such as "$\alpha$" and "$\Sigma$".

\begin{table}[h]
    \centering
    \resizebox{\linewidth}{!}{
    \begin{tabular}{lcccc}
        \toprule
        Symbol & GPT-4o & Claude 3.5 & Llama3.1-70B & Mistral-8x7B \\
        \midrule
        $\alpha$ & 25\% & 32\% & 45\% & 58\% \\
        $\beta$ & 22\% & 29\% & 38\% & 50\% \\
        $\Delta$ & 30\% & 35\% & 48\% & 60\% \\
        $\psi$ & 28\% & 33\% & 44\% & 57\% \\
        $\theta$ & 26\% & 31\% & 42\% & 55\% \\
        $\Sigma$ & 34\% & 38\% & 50\% & 63\% \\
        $\lambda$ & 20\% & 27\% & 36\% & 48\% \\
        $\pi$ & 18\% & 25\% & 34\% & 46\% \\
        $\mu$ & 24\% & 30\% & 41\% & 54\% \\
        $\xi$ & 27\% & 34\% & 46\% & 59\% \\
        \bottomrule
    \end{tabular}
    }
    \caption{Top 10 error-prone symbols and their misclassification rates.}
    \label{tab:error_symbols}
\end{table}

\begin{figure}[!h]
    \centering
    \includegraphics[width= \linewidth]{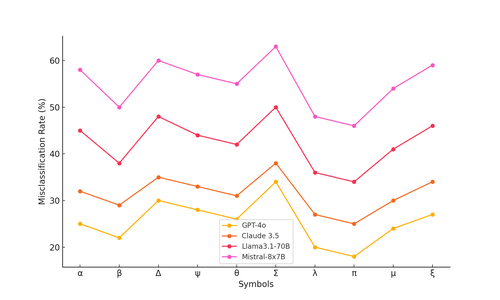}
    \caption{Misclassification rates for the top 10 error-prone symbols across different LLMs.}
    \label{fig:error_plot}
\end{figure}

\paragraph{Error Analysis by Symbol Type.}
The analysis revealed recurring error patterns: \textbf{(1) Ambiguity in Context}: Symbols like "$\alpha$" and "$\beta$" were frequently misclassified due to their diverse meanings in different disciplines. \textbf{(2) Overlapping Attributes}: "$\Sigma$" and "$\Delta$" were often confused between operators and constants depending on context. \textbf{(3) Domain-Specific Knowledge Gaps}: Models struggled with symbols like "$\psi$" and "$\theta$" that require understanding of physics or quantum mechanics.
\begin{table*}[!th]
    \centering
    \resizebox{\linewidth}{!}{
    \begin{tabular}{lll}
        \toprule
        Symbol & Contextual Sentence & Incorrect Prediction (Correct) \\
        \midrule
        $\alpha$ & "The coefficient $\alpha$ controls the rate of decay in the model." & Constant (Correct: Variable) \\
        $\psi$ & "The wave function $\psi$ represents the quantum state of the system." & Variable (Correct: Operator) \\
        $\Sigma$ & "The summation $\Sigma$ is over all possible states." & Operator (Correct: Constant) \\
        \bottomrule
    \end{tabular}
    }
    \caption{Qualitative analysis of errors for selected symbols.}
    \label{tab:qualitative_errors}
\end{table*}
\paragraph{Performance Comparison.}
Figure \ref{fig:error_plot} illustrates the misclassification rates for the top ten error-prone symbols across models. GPT-4o consistently outperformed smaller models, but the error rates remained substantial for challenging symbols.
\paragraph{Qualitative Examples.}
Table \ref{tab:qualitative_errors} provides examples of contextual sentences and the corresponding model predictions. These examples demonstrate how insufficient understanding of context leads to systematic errors.
Symbols like "\textit{$\alpha$}" and "\textit{$\delta$}" are often misclassified due to ambiguous contextual cues. For example, "\textit{$\alpha$}" is misclassified as a constant in differential equations, where it actually serves as a prefix operator given the context.

\paragraph{Impact of Context Length.}
While longer contexts improve classification accuracy overall, they also introduce noise. For example, in the full manuscript setting, models sometimes overfit on irrelevant context, leading to errors in identifying local operators as global constants.
 \\
 \\
\noindent
Through the case study, we conclude that despite the contextual advantages provided by STEM-POM, symbols with ambiguous or multifaceted roles remain challenging. Larger models like GPT-4o mitigate some of these issues but still fail to handle nuanced sub-attributes effectively.

\section{Discussions}
\label{appendix: discussion}
\paragraph{Impact and Applicability of STEM-PoM on Math Reasoning.}
In this section, we discuss why classifying mathematical tokens is highly relevant to advancing mathematical reasoning on LLMs.

On the surface level, summarization is a quintessential AI task, and being able to summarize mathematical tokens is a crucial downstream task that has already been held back in several recent research. One notable example is ScholarPhi \cite{head2021augmenting}, where researchers endeavor to make STEM manuscripts more accessible by enhancing a PDF viewer with an annotation tool that summarizes mathematical tokens and equations whenever users click on them. The authors observed that high-quality annotations significantly enhanced users' comprehension of the manuscripts. However, they manually annotated their test sets, as they were uncertain about how annotation accuracy would influence comprehension.

Furthermore, researchers have explored the task of converting mathematical content in manuscripts into computable functions \cite{greiner2022math, greiner2023making}. In such tasks, where mathematical expressions must be transformed into alternative formats, accurately classifying these tokens is crucial.

Therefore, the classification of mathematical tokens serves as a foundational component for numerous downstream tasks. We developed this dataset to provide researchers with a baseline for evaluating and advancing this critical functionality.

\section{Model Prompts}
\label{appendix: model_prompts}
\begin{figure*}[!ht]
\centering
\includegraphics[width=\textwidth]{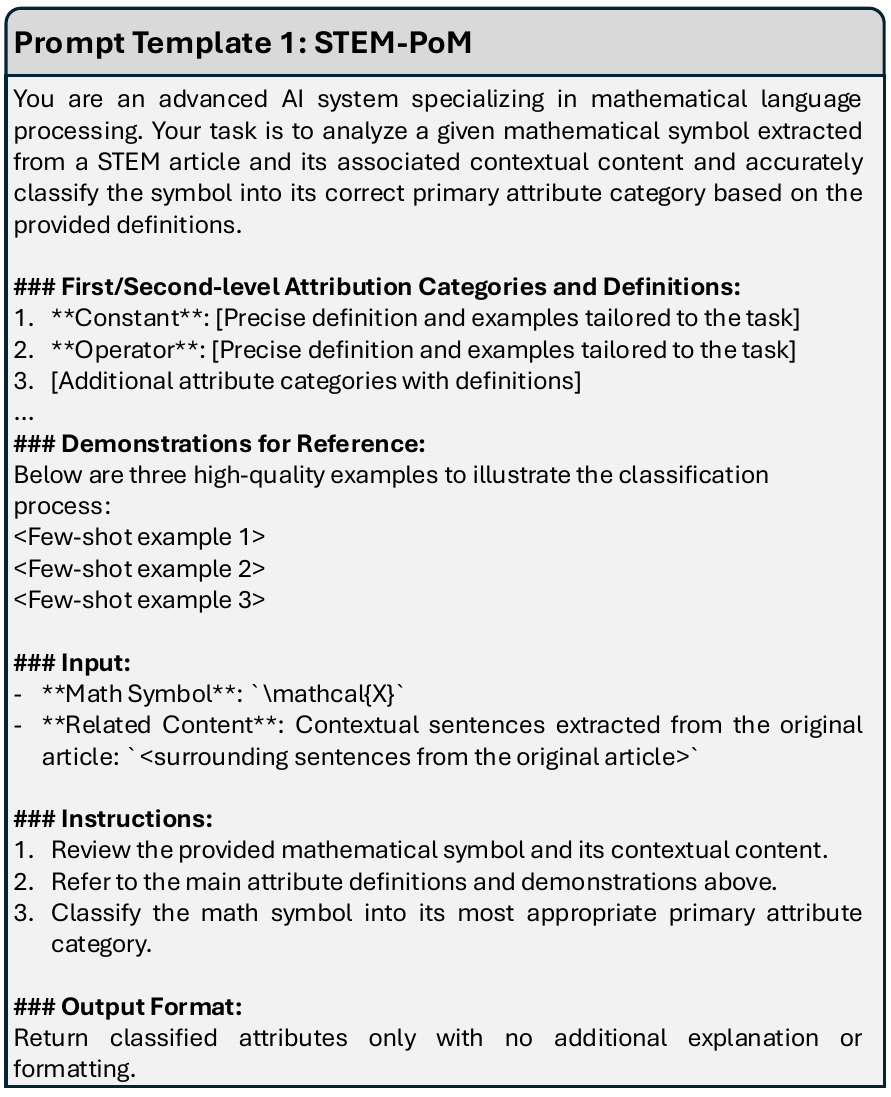}
\caption{The prompt design for \ours.}
\label{fig: stem_pom_prompt}
\end{figure*}

\begin{figure*}[!ht]
\centering
\includegraphics[width=\textwidth]{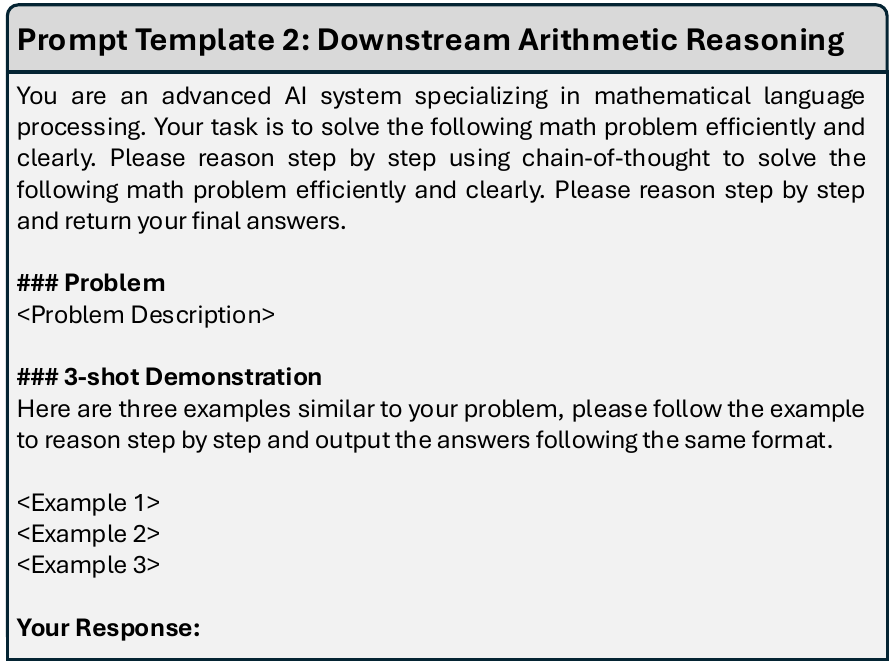}
\caption{The prompt design for downstream math reasoning tasks including GSM8K, MATH, and OlypiadBench.}
\label{fig: math_reasoning_prompt}
\end{figure*}

For detailed prompts of \ours and downstream math reasoning tasks, please refer to Figure \ref{fig: stem_pom_prompt} and \ref{fig: math_reasoning_prompt}.

\end{document}